\documentclass[runningheads]{llncs}

\usepackage{eccv}

\usepackage{eccvabbrv}

\usepackage{graphicx}
\usepackage{booktabs}

\usepackage[accsupp]{axessibility}  

\usepackage{hyperref}

\usepackage{orcidlink}

\begin{document}

\title{Few-Shot Image Classification and Segmentation as Visual Question Answering Using Vision-Language Models} 

\titlerunning{FS-CS as VQA Using VLMs}

\author{
    Tian Meng\inst{1,\dagger}\orcidlink{0000-0003-2330-2549} \and
    Yang Tao\inst{2,\dagger}\orcidlink{0000-0002-9625-2370} \and
    Ruilin Lyu\inst{1}\orcidlink{0009-0000-1026-6134} \and
    Wuliang Yin\inst{1,*}\orcidlink{0000-0001-5927-3052} 
    \\
    $^\dagger$Joint First Author\hspace{5em}
    $^*$Project Lead
}

\authorrunning{T.Meng et al.}

\institute{
    The University of Manchester \and 
    Mettler Toledo Safeline 
}

\maketitle

\begin{abstract}
    The task of few-shot image classification and segmentation (FS-CS) involves classifying and segmenting target objects in a query image, given only a few examples of the target classes. We introduce the Vision-Instructed Segmentation and Evaluation (VISE) method that transforms the FS-CS problem into the Visual Question Answering (VQA) problem, utilising Vision-Language Models (VLMs), and addresses it in a training-free manner. By enabling a VLM to interact with off-the-shelf vision models as tools, the proposed method is capable of classifying and segmenting target objects using only image-level labels. Specifically, chain-of-thought prompting and in-context learning guide the VLM to answer multiple-choice questions like a human; vision models such as YOLO and Segment Anything Model (SAM) assist the VLM in completing the task. The modular framework of the proposed method makes it easily extendable. Our approach achieves state-of-the-art performance on the Pascal-5i and COCO-20i datasets.

\end{abstract}    
\section{Introduction}
\label{sec:intro}
Few-shot image classification and segmentation (FS-CS) presents a formidable challenge in the field of computer vision, pertinent to tasks requiring both the identification of objects within an image and the precise delineation of their boundaries, all while operating under the constraint of extremely limited data for novel classes. Traditional methodologies \cite{kang_integrative_2022,kang_distilling_2023} to tackle FS-CS problems have predominantly revolved around intensive training regimes, often relying on meta-learning strategies \cite{hospedales_meta-learning_2020,nichol_first-order_2018,wang_meta-learning_2021} or transfer learning  \cite{weiss_survey_2016, zhu_transfer_2023, zhuang_comprehensive_2021} from broadly annotated base datasets to classes with sparse examples. While these conventional approaches have contributed valuable insights, they invariably encounter limitations, such as a propensity to overfitting, substantial computational expense, and the need for dataset-specific fine-tuning.

\begin{figure}[t]
  \centering
   \includegraphics[width=0.7\linewidth]{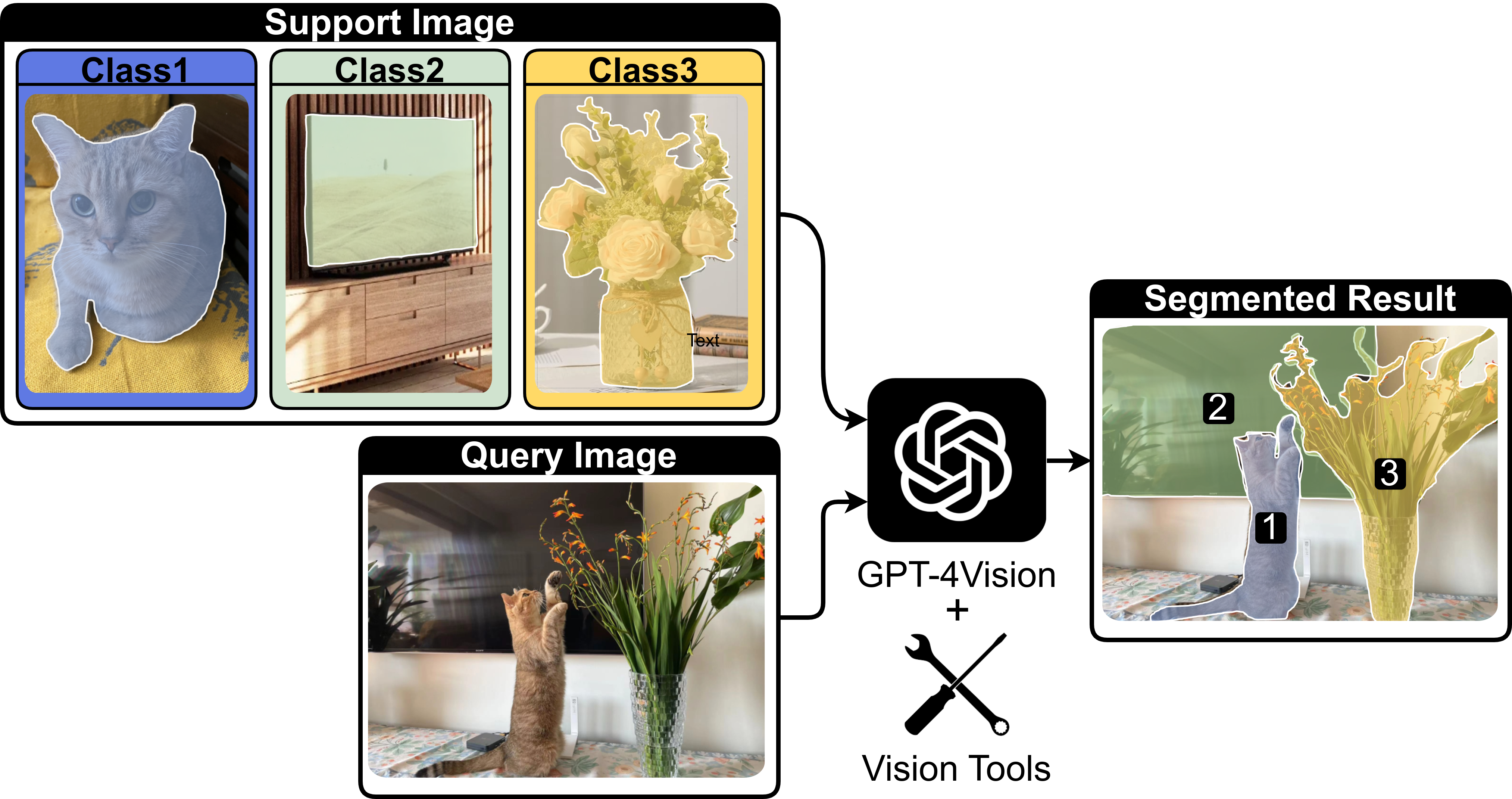}
   \caption{\textbf{Few-Shot Classification \& Segmentation Task Solved by Vision Language Models.} By providing vision tools to VLM like GPT-4Vision, it can solve the task of Few-Shot Image Classification and Segmentation with only image-level label in a training-free manner.}
   \label{fig:overview}
\end{figure}

Recent advancements in deep learning have spotlighted the capabilities of pretrained models, such as the Segment Anything Model (SAM) \cite{kirillov_segment_2023}, which has exhibited remarkable performance across various vision tasks without necessitating extensive retraining. A notable stride in this direction is the employment of visual prompting accompanied by human interaction, which enables these pretrained models to adapt their profound learned representations to specific tasks intuitively. This methodology allows for leveraging the generic, yet powerful, understanding these models have developed about visual data, circumventing the traditional training hurdles.

Parallel to this, the emergence of Vision-Language Models (VLMs) \cite{radford_learning_2021,ramesh_zero-shot_2021} has marked a significant evolution in handling complex visual problems through a multi-modal approach. VLMs combine the prowess of Large Language Models (LLMs) with an acute sense of visual understanding, making them adept at deciphering and analyzing tasks that combine both textual and visual cues. By interfacing directly with visual elements, VLMs extend beyond mere textual output, venturing into spatial and physical problem-solving realms. This capability opens a new vista for tasks like FS-CS, where the challenges involve more than just classifying and segmenting but also understanding the intricate visual semantics that interplay within images.

This work introduces a novel approach to FS-CS by leveraging the inherent strengths of VLMs, particularly in bypassing the necessity for traditional training paradigms and inherently employing visual prompting mechanisms. Herein, VLMs assume a central role, bridging the gap between high-level problem understanding and the execution of specific vision tasks akin to human-like reasoning and interaction with visual data. Through this method, we re-conceptualize FS-CS tasks as instances of Visual Question Answering (VQA), where the VLM, equipped with the context provided by few-shot examples, guides the process of image classification and segmentation. Our approach not only capitalizes on the advanced capabilities of VLMs in processing and understanding multi-modal data but also integrates seamlessly with state-of-the-art vision tools like YOLO and SAM, thereby crafting a symbiotic framework that excels at FS-CS tasks. This synergy, devoid of the traditional reliance on extensive training, showcases our method's robustness and adaptability in dealing with varied FS-CS challenges, establishing a new benchmark for efficiency and performance in the field.

To summarize, our contributions are the following:
\begin{itemize}



    \item We propose a training-free strategy, the Vision-Instructed Segmentation and Evaluation (VISE) modular framework, capitalizing on off-the-shelf vision tools and VLMs to navigate FS-CS challenges. This framework allows for swift adaptation across varied tasks and domains without intensive training.

    \item Our method redefines FS-CS as a VQA task, harnessing the reasoning powers of VLMs. Through visual and text prompting, we enable VLMs to interact with vision tools like YOLO and SAM, facilitating precise classification and segmentation with minimal supervision.

    \item Our approach sets new benchmarks on Pascal-5i \cite{shaban_one-shot_2017} and COCO-20i \cite{lin_microsoft_2014} datasets for FS-CS tasks. These advancements highlight the effectiveness of combining VLM reasoning and readily available vision tools within our flexible VISE framework.
\end{itemize}

\section{Related work}
\label{sec:relatedwork}

\subsection{Few-shot Image Classification and Segmentation}
FS-CS \cite{kang_integrative_2022,kang_distilling_2023} aims to generalize an algorithm to new classes not seen during training, given only a small sample of images. There are some previous studies navigating the challenges of minimal training data. Kang \textit{et al}. \cite{kang_distilling_2023} utilize Vision Transformers (ViTs) pre-trained through self-supervision, generating attention maps from ViT tokens to serve as pseudo-labels for FS-CS tasks, enhancing performance in mixed supervision settings. On the other hand, \cite{kang_integrative_2022} propose an integrative few-shot learning (iFSL) framework, combining classification and segmentation into a unified task. Their attentive squeeze network (ASNet) employs deep semantic correlation and global self-attention to produce accurate foreground maps from scarce examples. These works underscore a strategic move towards leveraging advanced representation and attention mechanisms, offering routes to tackle FS-CS challenges in a data-constrained environment. However, the generalization capability to new classes without further fine-tuning remains a substantial hurdle to overcome.


\subsection{Visual Question Answering}
VQA \cite{antol_vqa_2015,teney_visual_2017,wu_visual_2017,xue_variational_2023,yu_multi-level_2017} represents a pivotal intersection between computer vision and natural language processing, aiming to replicate comprehensive, human-like image understanding. The fusion of VQA into Vision-Language Models (VLMs) has unlocked groundbreaking possibilities across diverse applications. In robotics \cite{gao_physically_2024,saxena_multi-resolution_2023,brohan_rt-2_2023}, VLMs facilitate nuanced human-robot interaction, enabling robots to understand complex visual cues and verbal instructions, thus revolutionizing assistive technologies and automation. Similarly, in autonomous driving \cite{zhou_vision_2023}, VLM-integrated VQA systems enhance situational awareness, allowing vehicles to interpret dynamic road scenarios through a combination of visual observations and contextual questioning. Beyond these, VLMs are finding utility in advanced surveillance systems, where they can interpret and respond to visual and textual queries about activities and behaviors, bolstering security measures. These developments underscore VQA's transformative potential in elevating machine perception and interpretation to near-human levels, paving the way for innovative applications in technology-driven domains.

\section{Problem Formulation}
The core problem addressed in this paper is the few-shot image classification and segmentation, which involves both identifying the category of target objects in a query image and providing their precise segmentation mask, given only a handful of labeled examples. Formally, we consider the few-shot setting with a support set \( S \) and a query set \( Q \). For a \(N\)-way \(K\)-shot classification and segmentation task, the support set \( S \) consists of \(N\) distinct object classes, each with \(K\) labeled examples: 
\begin{equation}
  S = \{(x_{n,k}, y_{n,k}, m_{n,k})\}_{n=1,k=1}^{N,K}
  \label{eq:supportDef}
\end{equation}
where \(x_{n,k}\) is the \(k\)-th image instance of the \(n\)-th class, \(y_{n,k}\) is the class image-level label of image \(x_{n,k}\), and \(m_{n,k}\) is the binary mask for image \(x_{n,k}\), indicating the pixel-level presence of the \(n\)-th class object. The query set \( Q \) is defined as
\begin{equation}
  Q = \{(x', y', m')\} 
  \label{eq:queryDef}
\end{equation}
where \(x'\) is a new query image potentially containing instances of object classes from the support set \( S \), \(y'\) denotes the set of class labels present in \(x'\), and \(m'\) is the ground truth segmentation mask for the query image that needs to be predicted. The objective is to develop a model that uses the support set \( S \) to classify and segment the query image \(x'\) correctly.
\section{Method}
\label{sec:method}
\begin{figure*}
  \centering
  \includegraphics[width=1.0\linewidth]{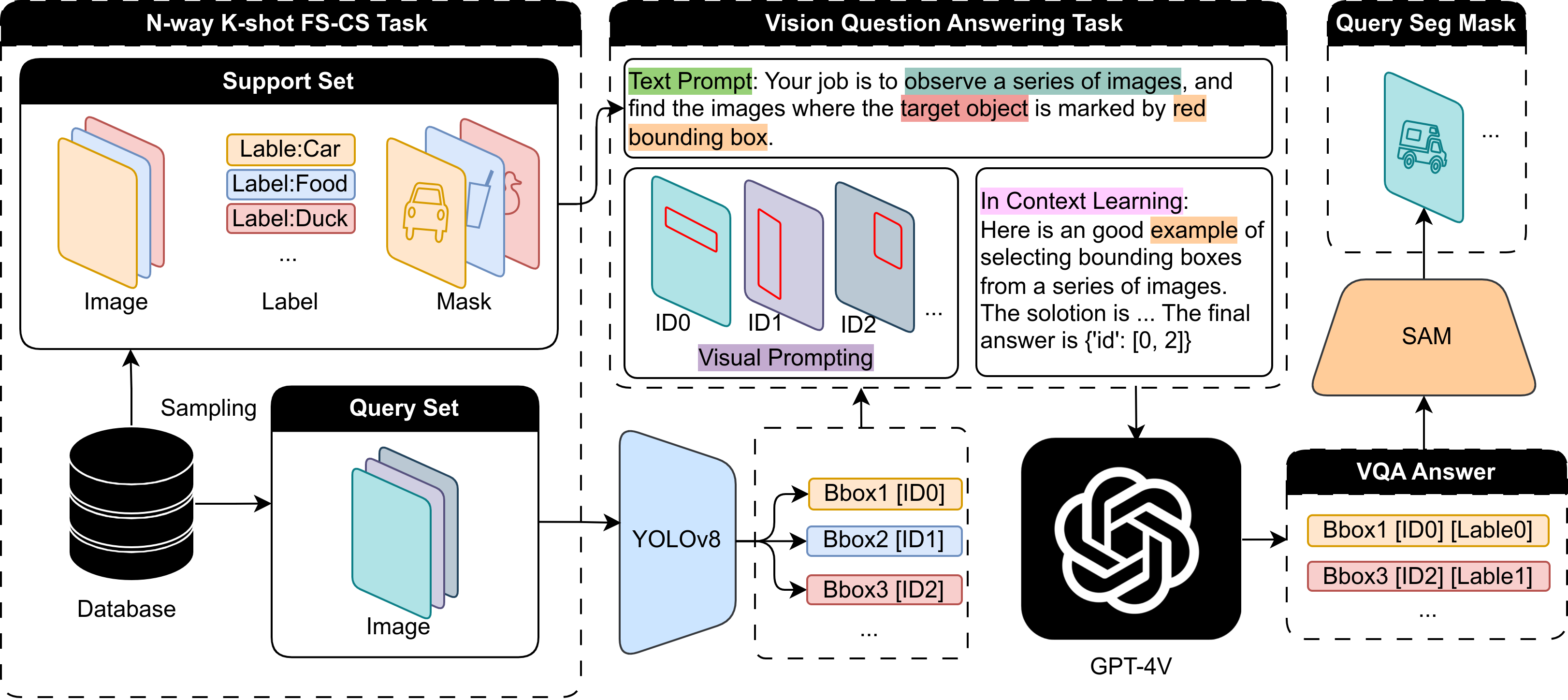}
  \hfill
  \caption{\textbf{VISE framework of VLM using visual tools to solve the FS-CS task.} First, a N-way K-shot FS-CS task is sampled from the database. Query images are given to object detection tool to get bounding boxes. Then, according to the support set, the original FS-CS task is transformed to a multi-choice VQA task. Last, image segmentation tool is used to obtain the ultimate segmentation mask of query set.}
  \label{fig:short}
\end{figure*}

Our approach to resolving the FS-CS problem reimagines it as a VQA task through the strategic utilization of VLMs. This section elaborates on the proposed method, including the creation of a modular framework, the formulation of FS-CS as a VQA problem, and the collaboration between VLM and vision models for task completion. Fig. \ref{fig:short} shows the overview of our framework.

\subsection{Overview of the Framework}
Our framework for solving the FS-CS task as VQA leverages the capabilities of VLMs and state-of-the-art vision tools. The process begins by sampling an N-way K-shot FS-CS task from a dataset. Each query image from this task is fed into an object detection model, such as YOLO, to obtain bounding boxes highlighting potential regions of interest. These bounding boxes serve as the basis for formulating the FS-CS task into a VQA paradigm, which effectively prompts the VLM to perform a multi-choice task contextualized by the support set.

Given the bounding boxes, the subsequent step involves the VLM in a VQA task that leverages both visual prompting and in-context learning. The VQA task is constructed as a series of questions asking the VLM to identify which bounding boxes enclose objects identical or similar to those in the support set. For clarity, let's define \(B = \{b_{id}\}_{id=1}^{M}\) as the set of bounding boxes identified in the query image \(x'\), with \(M\) being the total number of detected potential objects. The support set \(S\) is defined as in Eq. \ref{eq:supportDef}. The goal is to match each \(b_i\) to the correct class in \(S\), if applicable. The VQA task can thus be formulated as:

\textit{For each class \(n\) in the support set \(S\), identify bounding boxes \(b_i\) in \(B\) that enclose objects belonging to class \(n\).}

Applying visual prompting, we overlay the bounding boxes on the query image, and present this revised image alongside images from the support set \(S\) to the VLM, framing a multi-choice question. This setup utilizes the VLM's in-context learning capacity to comprehend the scenario semantically and select relevant bounding boxes.

Upon completing the classification through VQA, with each bounding box \(b_i\) associated with a predicted label \(\hat{y_i}\), the segmentation phase begins. We employ a vision tool, specifically the SAM, to perform precise segmentation within each \(b_i\). For each bounding box \(b_i\) that the VLM identifies as containing an object of interest, SAM generates a segmentation mask \(m_i\). Finally, we aggregate the segmentation masks of bounding boxes grouped by their assigned labels to construct the comprehensive class-specific segmentation masks for the query image:
\begin{equation}
M_n = \bigcup_{\substack{i=1 \\ \hat{y_i}=n}}^{M} m_i \quad \text{for each class } n \in \{1, 2, ..., N\}
\end{equation}
where \(M_n\) is the final segmentation mask for class \(n\), \(m_i\) is the segmentation mask for bounding box \(b_i\), \(M\) is the total number of bounding boxes, and \(y_i\) is the label assigned to the object within \(b_i\).

\subsection{VQA Task Formulation Details}
The formulation of the FS-CS problem as a VQA task intricately blends visual prompting with the in-context learning capabilities of VLMs. This synthesis not only contextualizes the task at hand for the VLM but also educates it to discern and segment objects in the query images with minimal direct supervision.

In the process of crafting a VQA task, the VLM is fed a series of structured multi-choice examples. These examples are designed to progressively refine the model's understanding and selection process. A critical aspect of this approach is the descriptive detail accompanying each visual prompt or bounding box. The descriptions aim to not only highlight the physical characteristics of the object within each bounding box but also to embed contextual clues that assist the VLM in making an informed classification.

For instance, each bounding box in the query image is described in detail, emphasizing distinctive features or attributes that correlate with the classes represented in the support set. Following these descriptions, the VLM is presented with a multi-choice question, challenging it to identify which of the visually prompted boxes contain objects belonging to the same classes as those in the support set. The structure of the VQA task would resemble the following pattern:

\begin{enumerate}
    \item \textit{Describe the object within a bounding box: "Here is a bounded area in the query image displaying a [brief description of the object based on shape, color, texture, etc.]." }
    \item \textit{Pose a multi-choice question: "Which of these objects is similar to the class exemplars provided in the support set?" Multiple options are provided, each corresponding to a bounding box with its description.}
\end{enumerate}

This method of direct description followed by a choice task significantly enhances the VLM's ability to pinpoint the correct images and bounding boxes, as it mimics a learning environment where examples are explicitly taught and reinforced through repetition. The in-context learning feature of the VLM thrives on this setup, as it effectively simulates an interactive learning scenario. Each example serves not only as a query but also as a lesson in distinguishing features relevant to the FS-CS task.

Through repeated exposure to these structured multi-choice examples, the VLM incrementally refines its ability to select the right bounding boxes. This approach inherently leverages the VLM's ability to digest complex instructions and learn from contextual cues, ensuring a more nuanced understanding of the task. The end goal is to prime the VLM to accurately classify and, subsequently, guide the segmentation of objects in the query images, drawing directly from the iterative learning experiences constructed within the VQA framework.

\subsection{Vision Tools for VLM Agent}

Leveraging state-of-the-art vision tools such as YOLOv8 for object detection and SAM for image segmentation transforms the VLM from a passive observer to an active participant in the FS-CS process. These tools act under the direction of the VLM, which uses the generated bounding boxes and subsequent segmentation masks to fulfill the FS-CS task without conventional training phases. Object detection with YOLOv8 provides an efficient and accurate means to identify potential objects of interest in the query images. Its role is crucial in the initial step, preparing the query for the VLM's comprehension and decision-making process.

Following the classification via the VQA framework, SAM's application for segmentation within the VLM-selected bounding boxes showcases how vision tools can extend the functionality of VLMs beyond simple yes/no or multiple-choice answers to actionable segmentation tasks. This collaborative interaction between VLMs and vision tools enables the flexible and effective application of pre-existing models to novel few-shot tasks, illustrating a significant advancement in leveraging cross-modal learning for practical computer vision challenges.

By leveraging this collaborative framework, our method can effectively address the FS-CS problem as a VQA task, benefiting from the nuanced understanding and adaptability of VLMs while utilizing the specialized capabilities of vision models for visual processing. A notable advantage of our method is its modular design, allowing for easy replacement or addition of vision models and VLMs. This flexibility ensures that the framework can adapt to advancements in model performance and task-specific requirements, facilitating continuous improvement and expansion to new domains or challenges within the FS-CS spectrum.
\section{Experiment}
In order to validate the efficacy of our proposed framework, we conducted extensive experiments on the widely used few-shot learning benchmark Pascal-5i and COCO-20i. We compared our model's performance against several state-of-the-art few-shot classification and segmentation approaches. This section details the experimental setup, datasets, evaluation metrics, baselines for comparison, and the results obtained.

\subsection{Datasets}
\textbf{Pascal-5i:} Derived from PASCAL VOC Challenge contains 20 object classes and is split into 4 different sets, with each set treated as one cross-validation fold. We follow the standard protocol for few-shot segmentation where each class is evaluated under the one-shot scenarios.\\\\
\textbf{COCO-20i:} Derived from the more challenging and diverse MS COCO dataset, it similarly divides the 80 object classes into 4 splits, with each containing 20 classes. The COCO-20i provides a more strenuous test of our method's effectiveness across a broader range of objects and scenarios.

\subsection{Metrics}
We follow the established evaluation protocol \cite{kang_integrative_2022,kang_distilling_2023} for each FS-CS benchmark.\\
\textbf{Classification Exact Ratio (ER):} The exact ratio is used to evaluate the performance of classification tasks, especially in a few-shot setting. It calculates the ratio of samples for which the predicted set of labels exactly matches the ground truth set of labels. 
\begin{equation}
    EMR = \frac{1}{|Q|} \sum_{(x', y') \in Q} \mathbb{1}[\hat{y}' = y']
\end{equation}
where \(Q\) denotes the query set, \(\hat{y}'\) is the predicted set of class labels for query image \(x'\), \(y'\) is the corresponding ground truth label set, and \(\mathbb{1}[\cdot]\) is the indicator function that is 1 if its argument is true and 0 otherwise. The EMR is expressed as a percentage, with higher values indicating better classification performance.
\\\\
\textbf{Segmentation Mean Intersection over Union (mIoU):} The mIoU serves as the primary metric for evaluating the segmentation aspect of FS-CS tasks. mIoU calculates the average of the ratio between the intersection and the union of the predicted and ground truth masks across all classes and query images, which is 
\begin{equation}
mIoU = \frac{1}{N} \sum_{n=1}^{N} \frac{\text{TP}_n}{\text{TP}_n + \text{FP}_n + \text{FN}_n}
\end{equation}
where \(N\) is the number of classes, \(\text{TP}_n\) is the number of true positive pixels for class \(n\) (i.e., pixels correctly identified as belonging to class \(n\)), \(\text{FP}_n\) is the number of false positive pixels (i.e., pixels wrongly identified as belonging to class \(n\)), and \(\text{FN}_n\) is the number of false negative pixels (i.e., pixels belonging to class \(n\) that were not identified). Higher mIoU values indicate better segmentation performance, with an mIoU of 100\% representing perfect segmentation.

\subsection{Implementation Details}
We employed publicly available pre-trained models of YOLOv8x for object detection and SAM VitH for segmentation due to their demonstrated effectiveness and efficiency. For the VLM, we utilized the GPT-4Vision from OpenAI. Our framework was implemented using LangChain and tested on an Nvidia RTX 4090 GPU. Throughout our experiments, we maintained a consistent in-context learning setup, dynamically adjusting prompts based on the task at hand. Bounding box proposals from YOLOv8x were filtered based on a confidence threshold of 0.5 to balance precision and recall.

\begin{table*}
  \centering
  \begin{tabular}{@{}lclclclclclclclclclc@{}}
\toprule
 &\multicolumn{10}{c}{1-way 1-shot}\\
 & \multicolumn{5}{c}{classification ER (\%)}&  \multicolumn{5}{c}{segmentation mIoU (\%)}\\
 Method& $5^0$& $5^1$& $5^2$& $5^3$& avg.& $5^0$& $5^1$& $5^2$& $5^3$&avg.\\
\midrule
    HSNet \cite{min_hypercorrelation_nodate}  & 84.5          & \textbf{84.8} & 60.8 & 85.3 & 78.9 & 20.0 & 23.5 & 16.2 & 16.6 & 19.1\\
    ASNet \cite{kang_integrative_2022}           & 80.2          & 84.0          & 66.2 & 82.7 & 78.3 & 11.7 & 21.1 & 13.4 & 16.2 & 15.6\\
    DINO  \cite{caron_emerging_2021}        & -             & -             & -    & -    & -    & 20.0 & 23.4 & 16.2 & 16.6 & 19.1\\
    CST   \cite{kang_distilling_2023}         & 84.0          & 82.2          & 70.8 & 82.6 & 79.9 & 35.8 & 38.9 & 28.9 & 29.2 & 33.2\\
    \textbf{VISE}                         & \textbf{92.3} & 81.5          & \textbf{86.7} & \textbf{91.5} & \textbf{88.0} & \textbf{61.9} & \textbf{55.3} & \textbf{56.7} & \textbf{57.5} & \textbf{57.9}\\
    \midrule
    &\multicolumn{10}{c}{2-way 1-shot}\\
    &\multicolumn{5}{c}{classification ER (\%)} & \multicolumn{5}{c}{segmentation mIoU (\%)}\\
    Method & $5^0$ & $5^1$ & $5^2$ & $5^3$ & avg. & $5^0$ & $5^1$ & $5^2$ & $5^3$ & avg.\\
    \midrule
    HSNet   & 70.8 & 67.0 & 36.5 & 70.1 & 61.1 & 11.0 & 23.4 & 15.4 & 17.0 & 17.7\\
    ASNet   & 67.6 & 70.2 & 44.5 & 69.3 & 62.9 & 10.3 & 20.7 & 12.6 & 15.9 & 14.9\\
    DINO    & -    & -    & -    & -    & -    & 20.0 & 23.4 & 16.2 & 16.6 & 19.1\\
    CST     & 74.3 & 67.2 & 49.0 & 67.8 & 64.6 & 35.7 & 36.0 & 26.8 & 29.1 & 31.9\\
   \textbf{VISE}    & \textbf{85.1} & \textbf{90.8} & \textbf{76.7} & \textbf{83.3} & \textbf{83.9} & \textbf{66.2} & \textbf{57.0} & \textbf{54.9} & \textbf{60.7} & \textbf{59.7}\\
    \bottomrule
  \end{tabular}
  \caption{Comparing model performance on Pascal-5i FS-CS of image-level setting.}
  \label{tab:pascal-1way1shot}
\end{table*}

\begin{table*}
    \centering
    \begin{tabular}{@{}lclclclclc@{}}
        \toprule
        \multicolumn{1}{c}{} & \multicolumn{2}{c}{1-way 1-shot} & \multicolumn{2}{c}{2-way 1-shot}\\
        Method          & ER (\%)    &  mIoU (\%)   &  ER (\%)     & mIoU (\%) \\
        \midrule
        DINO \cite{caron_emerging_2021}  & -    & 12.1 & -    & 7.4 \\
        CST  \cite{kang_distilling_2023}   & 78.2 & 19.6 & 62.4 & 18.3\\
        \textbf{VISE}            & \textbf{84.5} & \textbf{40.4} & \textbf{87.0} & \textbf{46.0}\\
        \bottomrule 
    \end{tabular}
    \caption{Comparing model performance on COCO-20i FS-CS of image-level setting.}
    \label{tab:coco}
\end{table*}

\subsection{Results and Evaluation}
Tables \ref{tab:pascal-1way1shot} summarize the performance of our approach on the Pascal-5i dataset for 1-way 1-shot and 2-way 1-shot settings, respectively. Our method outperforms existing state-of-the-art methods in segmentation mIoU, providing compelling evidence of its effectiveness in few-shot classification and segmentation tasks. It's noteworthy that our method achieved competitive performance in classification accuracy, showcasing its robustness across both aspects of the FS-CS problem. Similarly, Table \ref{tab:coco} shows our method's performance on the COCO-20i dataset, again emphasizing its superiority in handling complex segmentation tasks under a few-shot learning scenario. Despite the increased challenge posed by COCO-20i's diversity in object classes and backgrounds, our approach demonstrates remarkable results, particularly in the segmentation mIoU across both the 1-way and 2-way 1-shot settings.

Comparing to previous few-shot learning methods such as DINO \cite{caron_emerging_2021} and CST \cite{kang_distilling_2023}, the integration of VLM provides a significant performance boost, particularly in segmentation tasks, e.g., the mIoU for 2-way 1-shot task in COCO-20i is \(46.0 \%\). The ability to leverage extensive pre-trained knowledge and adapt it to specific few-shot contexts without extensive retraining underscores the potential of our method for practical applications. It is also worth mentioning that in scenarios where the contrast between the classes is starker, as in the 2-way 1-shot setting, our method exhibits even more pronounced improvement over competitors. This capability suggests that our approach effectively utilizes the provided few-shot examples to differentiate between classes, a crucial aspect of FS-CS tasks.

\begin{figure}[t]
  \centering
   \includegraphics[width=0.6\linewidth]{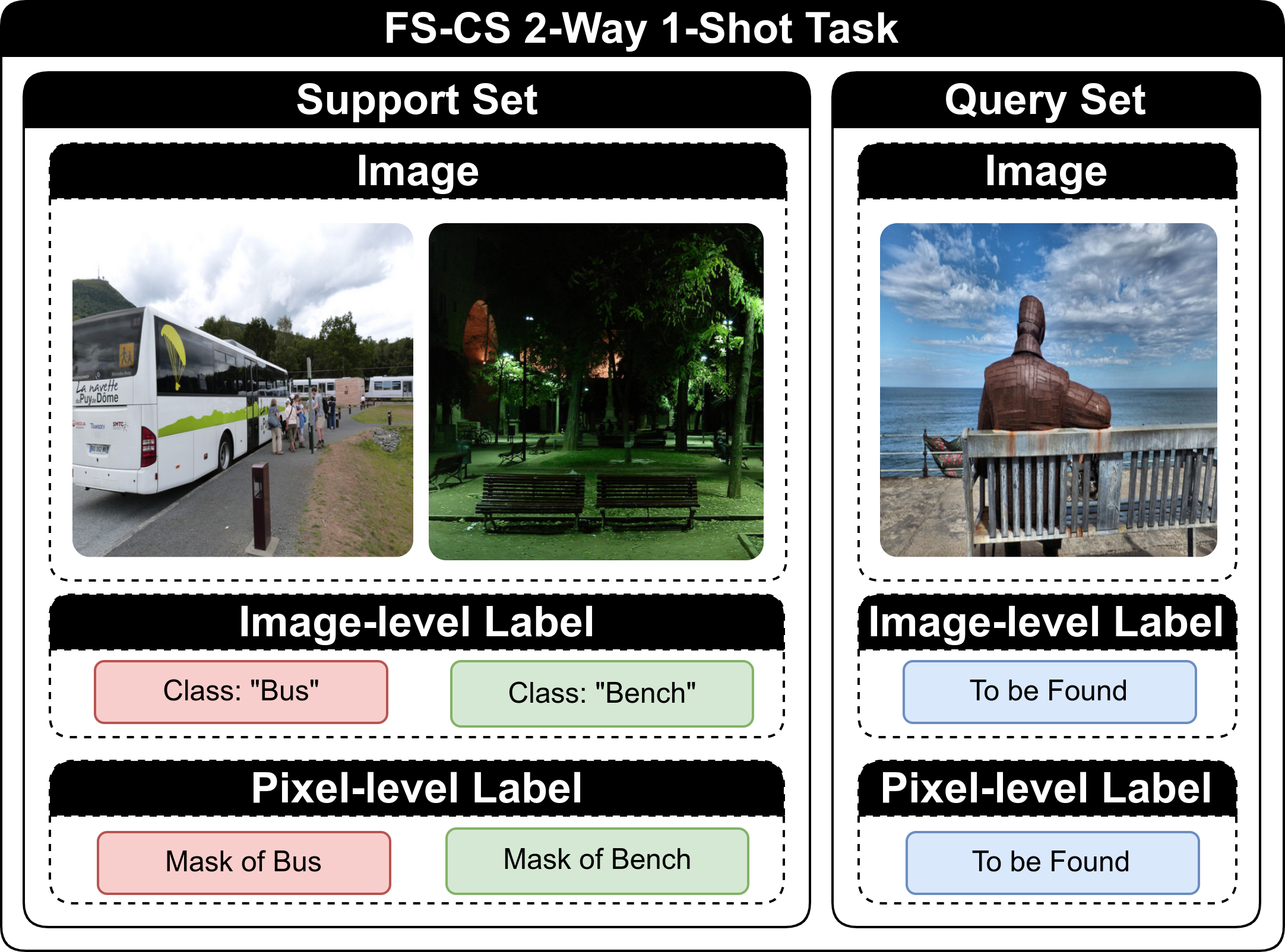}
   \caption{An example of 2-way 1-shot FS-CS task in COCO-20i Dataset}
   \label{fig:questing}
\end{figure}
\begin{figure}[t]
  \centering
   \includegraphics[width=1.0\linewidth]{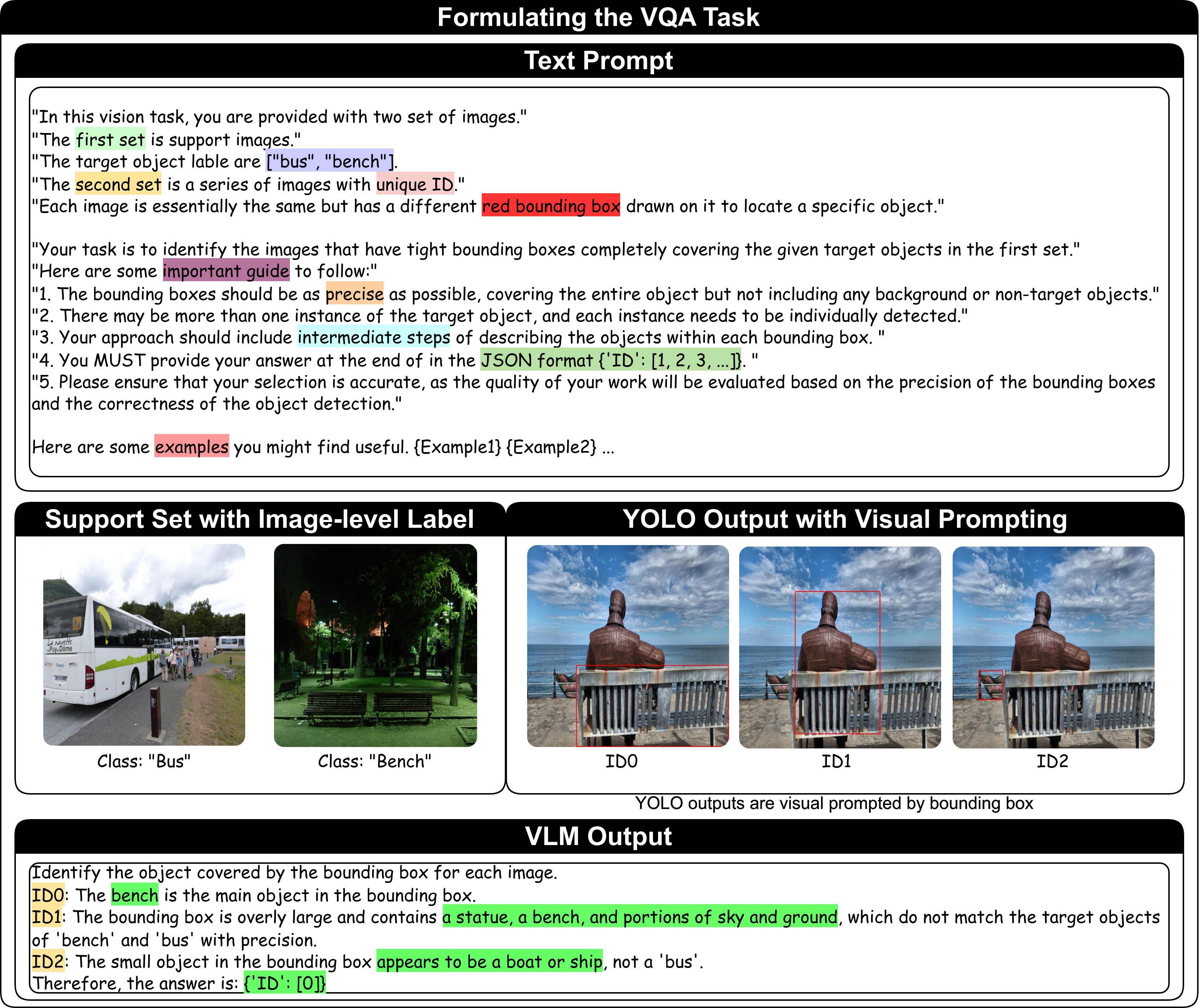}
   \caption{The VQA formulating for VLM.}
   \label{fig:questing}
\end{figure}

\subsection{Case Study}
To illustrate the efficacy and operational flow of our proposed approach, we meticulously detail the execution of one FS-CS task on a selected example from the COCO-20i dataset. This example not only showcases the methodological robustness of our framework but also provides insights into its practical application.\\
\textbf{Dataset Sampling.} From the COCO-20i dataset, a 2-way 1-shot FS-CS task was sampled, where the object classes were "Bus" and "Bench". The support set included: (a) One image of a bus and a bench. (b) The text label of these 2 images. (c) The corresponding binary mask indicating the pixel-level presence of each class. The query image sampled for this task depicted a seaside view featuring multiple objects, including a gray colored bench, a bronze statue of a human being and forward section of a boat. The challenge was to identify the presence of "bus" and "bench" objects within the query image and to generate accurate segmentation masks for each instance of these classes.\\
\textbf{Object Detection with YOLO.} The first step involved processing the query image through YOLOv8x to identify potential objects of interest. YOLOv8x, due to its efficacy and precision in object detection, returned bounding boxes for all detected objects within the image. These encompassed not only the target "bus" and "bench" classes but also various other objects present in the street scene.\\
\textbf{Formulating the VQA Task.} Given the bounding boxes determined by YOLOv8x, the next step was to transform the FS-CS challenge into a VQA paradigm. For each detected object in the query image, we constructed a descriptive prompt that encapsulated the visual characteristics discernible from its bounding box. The VLM, leveraging its in-context learning capabilities, then discerned whether the highlighted objects corresponded to either of the "Bus" or "Bench" classes as presented in the support set.\\
\textbf{Segmentation with SAM.} Upon identifying the relevant bounding boxes containing instances of the "Bus" and "Bench" classes, the final step was to generate precise segmentation masks for these objects. This task was conducted by SAM, a segmentation tool known for its precision and adaptability. SAM processed each identified bounding box independently, meticulously generating binary masks that demarcated the exact pixel-level presence of the target objects—ensuring a high degree of accuracy in segmentation output for both classes within the query image.\\
This example illustrates the workflow and the nuanced execution of our FS-CS methodology. By transforming the FS-CS task into a VQA challenge and leveraging the synergistic capabilities of YOLOv8x and SAM under the orchestration of a VLM, our method demonstrates a compelling approach to few-shot learning, significantly advancing the state-of-the-art in visual question answering paradigms. More examples including the under-performing tasks are included in the supplement materials.

\subsection{Ablation Experiment}
To assess the significance of integrating specialized vision tools in our framework, we conducted an ablation study focusing on the roles of 2 vision tools in the overall performance of FS-CS tasks. In our proposed method, we utilize YOLOv8x, a state-of-the-art object detection model, to identify regions of interest within query images, and SAM to obtrain segmentation mask for each class. However, to explore the impact of replacing these vision tools with the capabilities inherent in VLMs, we prompted a VLM to perform object detection and segmentation directly, bypassing the use of YOLOv8x and SAM. This experiment aimed to understand the importance of using dedicated vision tools for the initial detection phase and last segment generation in our problem-solving framework. For this purpose, we employed GPT-4V to generate bounding boxes in string format instead of relying on any object detection tool. The VLM was prompted to identify objects within the query images and produce textual representations of their bounding boxes. Subsequently, these bounding box descriptions were parsed and utilized to direct the SAM in generating segmentation masks, following the same workflow outlined in our primary approach. In addition, we also ask VLM to generate True or False segmentation masks directly, bypassing the use SAM.\\
The results of this ablation experiment are encapsulated in the Tables \ref{tab:ablation}. GPT4V couldn't generate segmentation stably and effectively, which results in bad results that can't be statistically accounted for. In addition, It became evident that the replacement of YOLOv8x with the VLM-generated bounding boxes led to a noticeable degradation in the performance metrics, particularly in the accuracy of segmentation (mIoU) across both datasets. This decline can primarily be attributed to the inaccuracies and uncertainties associated with bounding boxes generated by the VLM. The string-based bounding boxes lacked the precision and reliability of those produced by a dedicated object detection model like YOLOv8x, leading to sub-optimal inputs for the SAM segmentation tool.

\begin{table*}
    \centering
    \begin{tabular}{@{}lclclclclclclclclclc@{}}
        \toprule
        &\multicolumn{10}{c}{1-way 1-shot}\\
        &\multicolumn{5}{c}{classification ER (\%)}&  \multicolumn{5}{c}{segmentation mIoU (\%)}\\
        Method& $5^0$& $5^1$& $5^2$& $5^3$& avg.& $5^0$& $5^1$& $5^2$& $5^3$&avg.\\
        \midrule
        GPT4V           & 78.9 & \textbf{83.4} & 90.6 & \textbf{91.2} & \textbf{86.0} & -    & -    & -    & -    & -   \\
        GPT4V+SAM       & \textbf{83.2} & 80.9 & 87.6 & 88.7 & 85.1 & 9.3  & 6.4  & 7.2  & 17.5 & 10.1\\
        YOLO+GPT4V+SAM  & 80.3 & 79.5 & \textbf{91.4} & 86.7 & 84.5 & \textbf{43.1} & \textbf{24.7} & \textbf{42.5} & \textbf{49.4} & \textbf{40.4}\\
        \midrule
        &\multicolumn{10}{c}{2-way 1-shot}\\
        &\multicolumn{5}{c}{classification ER (\%)}&  \multicolumn{5}{c}{segmentation mIoU (\%)}\\
        Method& $5^0$& $5^1$& $5^2$& $5^3$& avg.& $5^0$& $5^1$& $5^2$& $5^3$&avg.\\
        \midrule
        GPT4V           & 80.2 & 81.3 & \textbf{95.6} & \textbf{93.8 }& \textbf{87.7} & -    & -    & -    & -    & -   \\
        GPT4V+SAM       & 79.5 & \textbf{84.7} & 92.0 & 90.6 & 86.7 & 20.1 & 12.6 & 21.2 & 22.9 & 19.2\\
        YOLO+GPT4V+SAM  & \textbf{82.6} & 83.0 & 93.6 & 88.9 & 87.0 & \textbf{37.0} & \textbf{42.9} & \textbf{46.9} & \textbf{57.3} & \textbf{46.0}\\
        \bottomrule
        \end{tabular}
    \caption{Ablation study comparison of using vision tools in the FS-CS framework on COCO-20i.}
    \label{tab:ablation}
\end{table*}

The comparative results consolidated in Table \ref{tab:ablation} reinforce the critical role of utilizing dedicated vision tools, such as YOLOv8 and SAM, for object detection within our FS-CS solution framework. While VLMs like GPT-4Vision possess remarkable capabilities in terms of in-context learning and understanding complex tasks, their direct application in generating bounding boxes for object detection and mask for semantic segmentation exhibits significant limitations when compared to outputs from specialized vision models.

This ablation study underscores a pivotal insight: the integration of VLMs with high-performance vision tools offers a synergistic effect that significantly elevates the accuracy and reliability of FS-CS tasks. The precision in object detection provided by models like YOLOv8 ensures that subsequent stages of processing, such as segmentation by SAM, are based on accurate and reliable inputs, thereby enhancing overall task performance. The experiment distinctly highlights that for achieving state-of-the-art results in few-shot classification and segmentation, leveraging the strengths of both VLMs for their reasoning and natural language processing capabilities, and dedicated vision models for their high-precision visual processing, is indispensable.
\section{Conclusion}
In conclusion, the experimental results across two benchmark datasets validate the effectiveness of our proposed framework in addressing few-shot classification and segmentation tasks. By leveraging the capabilities of VLMs and state-of-the-art vision tools, our approach not only simplifies the learning process but also achieves superior performance, especially in segmentation tasks. This establishes a strong case for further exploration and development of VLM-based methods in few-shot learning domains.

%
%
\bibliographystyle{splncs04}
\bibliography{main}

\clearpage
\setcounter{page}{1}

\section{Supplementary Materials}
\label{sec:supplementary}
This supplementary material provides deeper insights into the our VISE method, expanding upon the successes and limitations encountered in our experimentation. We offer a detailed analysis of instances where our framework did not perform as expected, underscoring the critical aspects leading to classification mistakes and low mIoU scores. Additionally, we present a set of cases that illustrate the robustness and accuracy of our method, serving to highlight the potential of leveraging VLMs in conjunction with state-of-the-art vision tools for FS-CS tasks.

\subsection{A1: Classification mistakes}
\begin{figure}[t]
  \centering
   \includegraphics[width=0.9\linewidth]{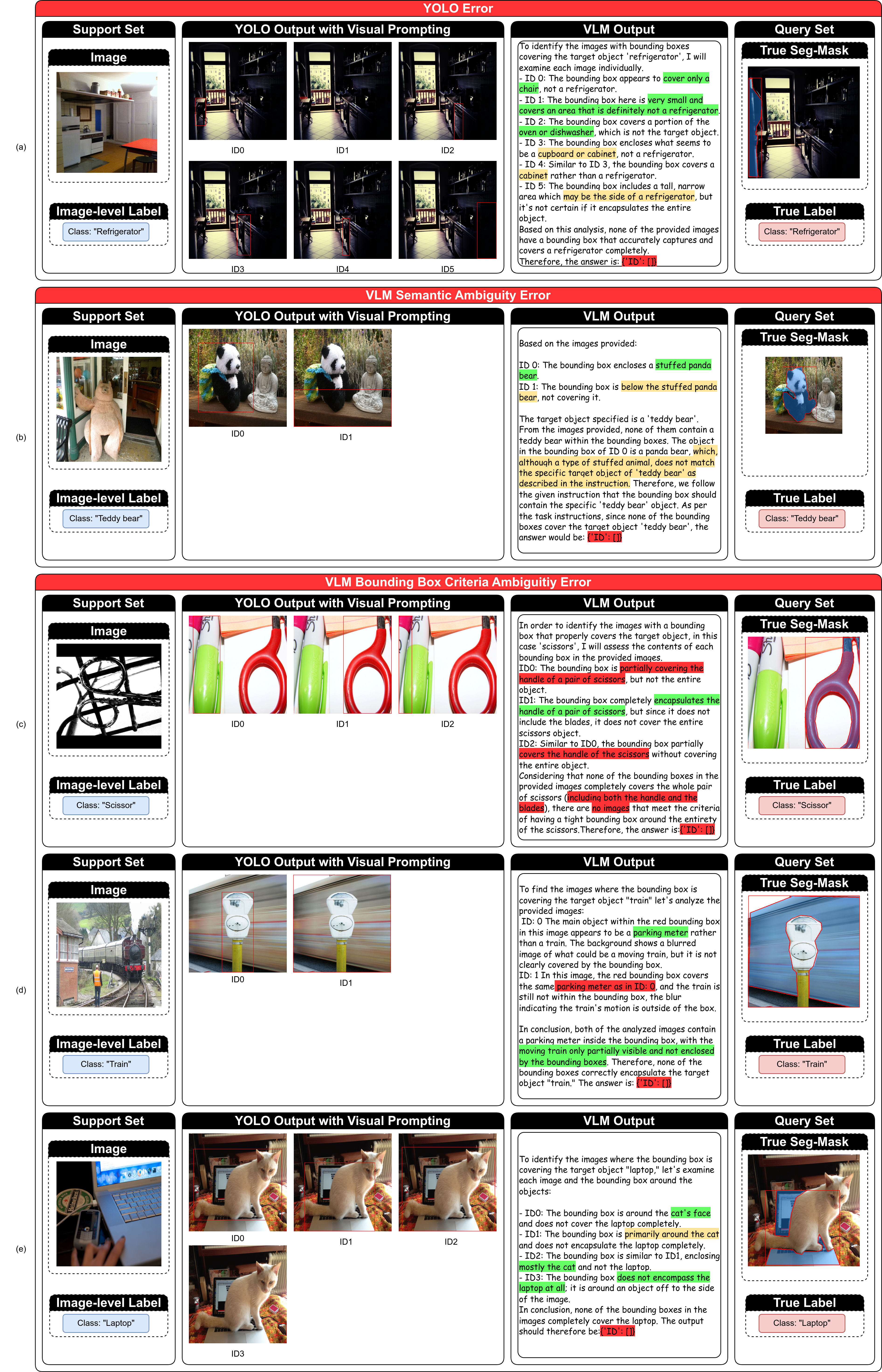}
   \caption{Classification mistakes. The location of error is marked in red, ambiguous conclusion is marked in yellow, and correct result is marked in green.}
   \label{fig:classification_mistake}
\end{figure}

Error analysis is crucial for understanding the limitations of our proposed method and for guiding future improvements. Classification mistakes generally fell into two main categories: errors caused by the object detection model (YOLO) and errors originated from the VLM's processing (see Fig. \ref{fig:classification_mistake}). \\\\
\textbf{YOLO Errors:} The first example (a) in Fig. \ref{fig:classification_mistake} showcases an instance where YOLO did not provide a correct bounding box for the refrigerator. The object's location at the periphery of the image, combined with inadequate lighting, significantly contributed to this miss. Such errors underscore the challenge of accurately detecting objects under varied environmental conditions.\\\\
\textbf{VLM Errors:} Errors from VLM processing can be categorized into semantic ambiguities and bounding box criteria ambiguities. In the case of semantic ambiguity, as illustrated by the (b) in the figure, a teddy bear designed to resemble a panda was correctly identified, yet it was ultimately not selected for classification. This misstep highlights dilemmas in object classification where boundary definitions of classes may overlap or appear ambiguous to the model. Bounding box criteria ambiguities arose when either an overly inclusive or overly tight bounding box was proposed by YOLO, affecting the VLM's classification decision. Several examples (c), (d) and (e) in Fig. \ref{fig:classification_mistake} depict how bounding boxes that encompass multiple objects or insufficiently capture an object's entirety can mislead the VLM and complicate accurate classification.

\subsection{A2: Low mIoU cases}
\begin{figure}[t]
  \centering
   \includegraphics[width=1.0\linewidth]{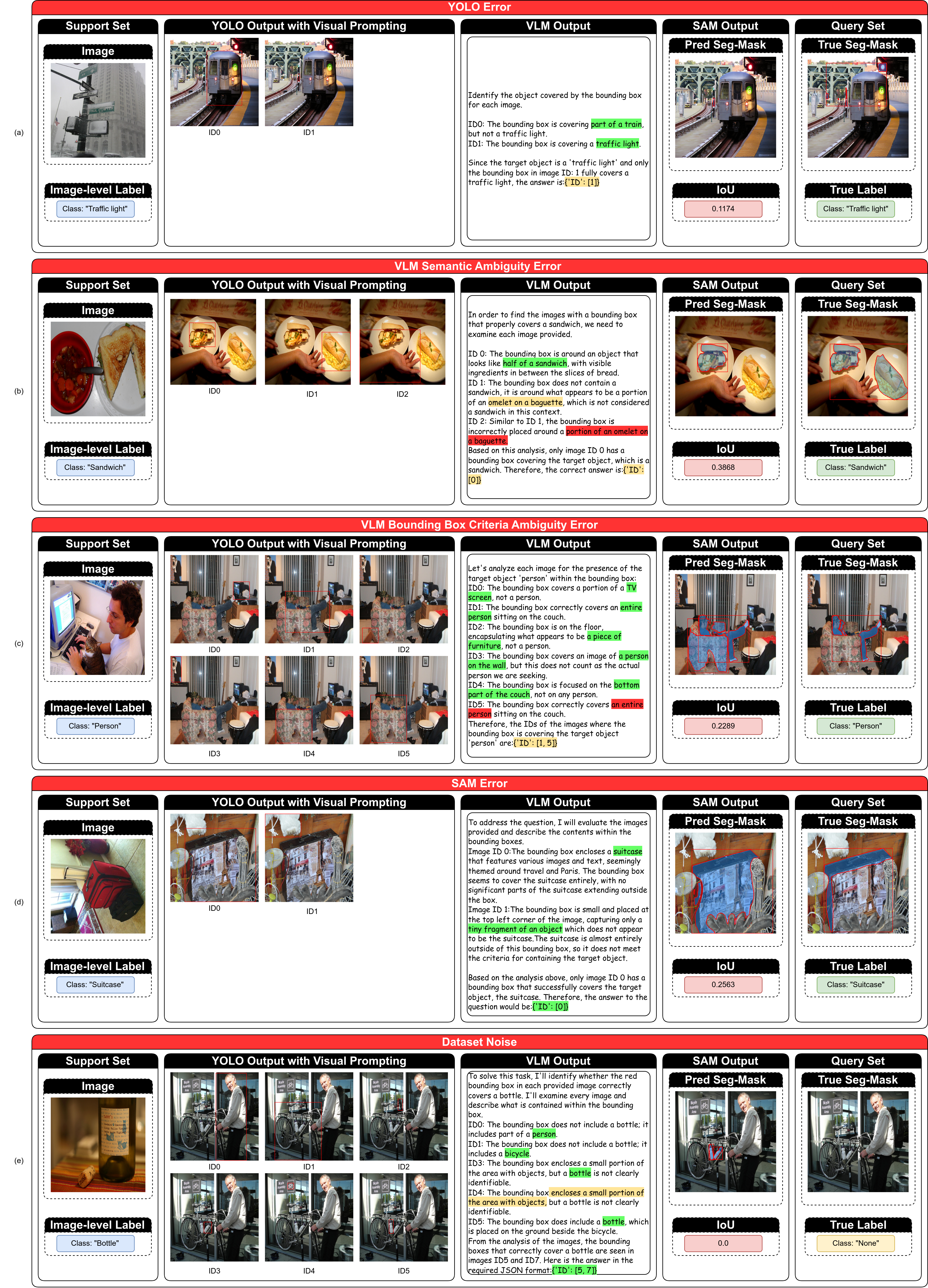}
   \caption{Low IoU cases. The location of error is marked in red, ambiguous conclusion is marked in yellow, and correct result is marked in green.}
   \label{fig:low_iou}
\end{figure}

Analyzing cases of low mIoU is instrumental in pinpointing specific errors across the three primary stages of our method: YOLO, VLM, and SAM errors, along with instances of dataset noise (see Fig. \ref{fig:low_iou}).\\\\
\textbf{YOLO Errors:} Small objects pose a significant challenge, as demonstrated by a traffic light in Fig. \ref{fig:low_iou} (a) not being accurately bounded due to its size. Such misses suggest a limitation in detecting objects that occupy minimal space within the image frame.\\\\
\textbf{VLM Errors:} Similar to the classification mistakes, errors in segmentation often stem from semantic ambiguities and bounding box criteria ambiguities. The illustrated case (b) of a sandwich mislabelled as an omelet exemplifies semantic ambiguity, revealing difficulties in precise classification when objects share visual or contextual similarities. Bounding box criteria ambiguity affects segmentation when a box includes multiple objects or does not fully encapsulate the target object. An instance of a person sitting on a sofa in Fig. \ref{fig:low_iou} (c), with the bounding box including both entities, resulted in an erroneous selection by the VLM.\\\\
\textbf{SAM Errors:} Although SAM generally performed well in generating segmentation masks, inaccuracies occurred in certain cases. For instance, a suitcase in Fig. \ref{fig:low_iou} (d) identified by the system had an imperfectly generated mask, indicating challenges in delineating complex object shapes accurately.\\\\
\textbf{Dataset Noise:} An additional obstacle in achieving high mIoU scores was noise within the dataset itself as shown in Fig. \ref{fig:low_iou} (e). Instances where our method successfully identified objects not labeled in the dataset underscore discrepancies between ground truth annotations and actual image content.

\subsection{A3: Good cases highlight}
\begin{figure}[t]
  \centering
   \includegraphics[width=1.0\linewidth]{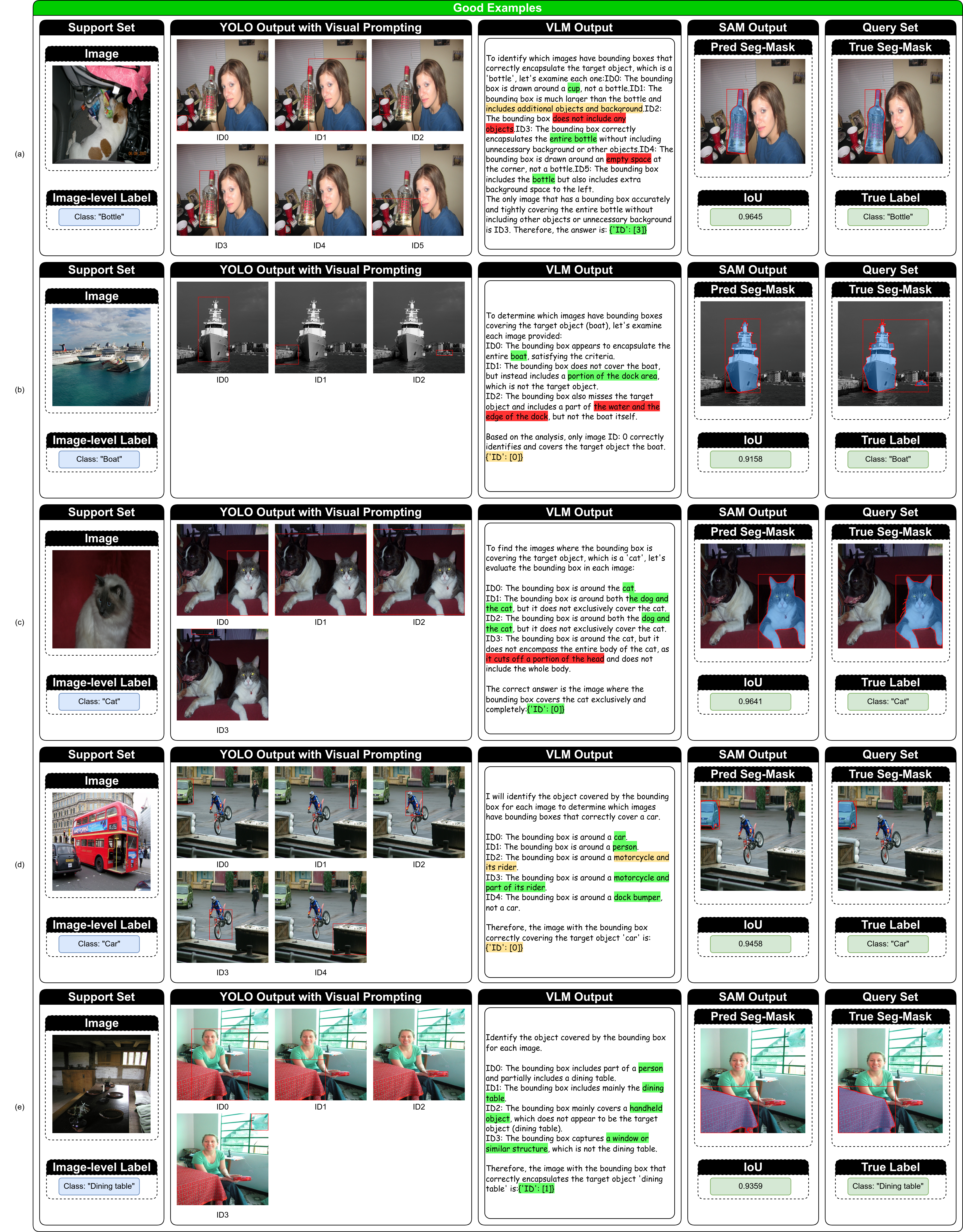}
   \caption{Some good cases with high IoU. The location of error is marked in red, ambiguous conclusion is marked in yellow, and correct result is marked in green.}
   \label{fig:good}
\end{figure}

Despite the challenges and errors observed, the VISE method demonstrated considerable strength in tackling FS-CS tasks through a novel VQA approach. Fig. \ref{fig:good} showcases examples where our method achieved high IoU scores, highlighting the effectiveness and adaptability of integrating VLMs with advanced vision tools in accurately classifying and segmenting objects given limited examples. Our analyses of both successful cases and limitations enrich our understanding of the VISE method's current capabilities and pave the way for future enhancements. By addressing identified errors and further refining the interplay between vision-language models and vision tools, we anticipate substantial advancements in few-shot image classification and segmentation performance.

\end{document}